\documentclass{article}

\PassOptionsToPackage{numbers,compress}{natbib}
\usepackage[preprint]{neurips_2026}

\usepackage[utf8]{inputenc}     
\usepackage[T1]{fontenc}        
\usepackage{url}                
\usepackage{booktabs}           
\usepackage{multirow}           
\usepackage{amsfonts}           
\usepackage{amsmath}            
\usepackage{nicefrac}           
\usepackage{microtype}          
\usepackage{xcolor}             
\usepackage{tikz}               
\usepackage{pgfplots}
\pgfplotsset{compat=1.18}
\usepackage{algorithm}
\usepackage{algorithmic}
\usepackage{placeins}
\usepackage{hyperref}
\hypersetup{colorlinks=true, citecolor=blue, linkcolor=blue, urlcolor=blue,
  pdftitle={When Simpler Beats Learned: Dual-Weighted Sampling vs. Machine Learning for the Pricing Subproblem in Graph Coloring},
  pdfauthor={Nikolai Antonov, Přemysl Šůcha, Mikoláš Janota, Jan Hůla}}

\title{When Simpler Beats Learned: Dual-Weighted Sampling vs.\ Machine Learning for the Pricing Subproblem in Graph Coloring}

\author{%
  Nikolai Antonov \quad Přemysl Šůcha \quad Mikoláš Janota \quad Jan Hůla\\
  Czech Technical University in Prague\\
  \texttt{antonni1@fel.cvut.cz}
}

\begin{document}

\maketitle

\begin{abstract}
The graph coloring problem can be solved to optimality by branch-and-price (B\&P), but its performance hinges on the column generation (CG) step, where each iteration requires solving a maximum weight independent set (MWIS) pricing problem.
The state-of-the-art machine-learning (ML) pricing heuristic follows a prediction--sampling paradigm: a learned model scores each vertex of a graph, and independent sets are sampled from the induced distribution.
We observe, however, that the scoring procedure does not need to be learned, and critically assess the role of ML by replacing the learned scores with the dual solution of the restricted master problem, resulting in Dual-Weighted Sampling --- a proposed lightweight alternative that requires no feature engineering or training.
Within a unified CG and B\&P framework that varies only the pricing oracle, Dual-Weighted Sampling outperforms MLPH, hand-crafted heuristics, and exact solvers on the established graph coloring benchmark in both runtime and solution quality. In particular, it remains competitive on the DIMACS random family --- where MLPH was shown strongest in the original evaluation --- while achieving more than a two-fold speedup in CG runtime on recursive-sparse and real-world graph families.

\end{abstract}

\section{Introduction}
\label{sec:introduction}

Branch-and-price (B\&P) is among the most effective exact methods for combinatorial
optimization problems~\cite{Barnhart1998}. Its core idea is to solve the LP relaxation
of the problem via column generation (CG)~\cite{LubbeckeDesrosiers2005} at every node
of a branch-and-bound search tree. The central computational challenge within CG is the \emph{pricing subproblem}: at each iteration, the algorithm must identify a column with negative reduced cost, which, in the graph coloring setting, corresponds to solving a maximum weight independent set (MWIS) problem. Since exact MWIS is NP-hard, each pricing call is a potential bottleneck, and practical B\&P implementations rely on heuristic pricing to remain tractable.

An active line of research related to heuristics aims to integrate machine learning (ML) into B\&P, targeting branching decisions, node selection, and pricing-step
acceleration~\cite{Bengio2021,GasseNeurIPS2019,Morabit2021,ChiNeurIPS2022,Vclavk2018,Kouteck2025}. The common premise is that a learned model can capture structural regularities difficult to encode explicitly. In this vein, Shen et al.~\cite{Shen2022} propose a machine-learning pricing heuristic (MLPH) that follows a \emph{prediction--sampling} paradigm: a linear support vector machine (SVM) trained on features derived from the dual solution and graph structure assigns each vertex a score reflecting its likelihood of belonging to a high-quality independent set, and candidate independent sets are then sampled from the induced distribution. MLPH
represents the state of the art among heuristic pricing methods for graph coloring via B\&P and serves as the primary baseline in our study.

We note, however, that the ML-based enhancements carry a non-negligible cost: this concerns collecting training data and feature engineering, while the trained models can degrade under distribution shift across instance families. Looking more closely at the prediction--sampling paradigm, we observe that \emph{the scoring rule need not be learned}, and ask whether a simpler, training-free mechanism can provide equally good---or even better---guidance. Our central contribution is to take the dual solution of the restricted master problem (RMP) directly as the vertex scores, feeding them into the same sampling procedure that MLPH uses; we call this approach \emph{Dual-Weighted Sampling}, abbreviated \emph{Dual}. Notably, Dual uses only the RMP dual solution as input and requires no training and no feature engineering. Prior work comparing exact solvers and handcrafted heuristics has left this baseline unexamined.

Our paper fills this gap. We embed seven pricing oracles---Dual, MLPH, uniform Random sampling, ant colony optimization (ACO), fast weighted-clique local search (FastWClq), integer linear programming (ILP), and constraint programming (CP)---into a unified CG and B\&P framework while holding all other components fixed, so that observed differences in performance can be attributed solely to the pricing step. The evaluation runs on the established Google graph coloring benchmark, with results analyzed separately across four structural macro-families. Our contributions are as follows:

\begin{itemize}
    \item \textbf{Dual-Weighted Sampling attains the best aggregate performance among the evaluated pricing oracles.} Within the unified framework, Dual achieves the lowest CG runtime and the lowest mean RMP objective on every macro-family of the Google benchmark. A face-to-face per-family analysis of Dual versus MLPH shows that Dual is consistently faster: the largest CG-runtime gains concentrate on the recursive-sparse ($4.4\times$) and real-world ($2.8\times$) families; on the DIMACS random family---where MLPH was reported strongest in the original evaluation---the gap shrinks to $1.5\times$ but Dual still leads. The advantage, though in milder form, carries over to branch-and-price.

    \item \textbf{Learned scoring does not provide additional benefits.} MLPH applies a trained SVM on top of a set of features derived from the dual solution and graph statistics; Dual drops that SVM and feeds the dual solution directly into the same sampling procedure. Since Dual outperforms MLPH on every benchmark family, we conclude that the additional features and SVM that constitute MLPH provide no benefit beyond using the dual solution itself.
\end{itemize}

The rest of the paper is organized as follows. Section~\ref{sec:related_work} reviews related work. Section~\ref{sec:problem} formalizes the graph coloring problem, column generation, and the MWIS pricing subproblem. Section~\ref{sec:methodology} describes the seven pricing oracles and the unified CG and B\&P framework. Experimental results and the per-family Dual--MLPH comparison are presented in Section~\ref{sec:experiments}. Section~\ref{sec:conclusion} concludes the work.
\section{Related Work}
\label{sec:related_work}

Branch-and-price~\cite{MehrotraTrick1996,LubbeckeDesrosiers2005,Lewis2021} is among the most effective exact methods for combinatorial optimization, and its methodology has been continually refined along several research directions. The most established approaches focus on stabilizing the restricted master problem~\cite{Malaguti2011}, deriving tighter lower bounds on the chromatic number~\cite{Held2012}, developing more advantageous branching rules~\cite{Morrison2014} and accelerating solving of the pricing subproblem~\cite{GualandiMalucelli2012}. We focus on the last direction shaped by an inherent speed--quality trade-off: exact pricing finds the column with the highest improving potential, but it is computationally expensive to use it; heuristic pricers can provide improving columns of a good quality at much cheaper cost, which makes them relevant for practical B\&P implementations.

Building on the practical importance of heuristic pricing, machine learning has been introduced into CG and B\&P at several decision points: learning branching policies~\cite{GasseNeurIPS2019}, policies for selecting the most valuable columns from the generated pool~\cite{ChiNeurIPS2022}, and learning to generate initial columns so as to reduce the number of CG iterations in the main loop~\cite{SunErnst2023}. Machine learning has also been applied to predict graph colorings directly~\cite{DasAhmad2019,CumminsVeras2024}, though these methods lie outside the CG setting. Closest to our topic is the use of ML for heuristic solving of the pricing subproblem~\cite{Shen2022}, which instantiates the \emph{prediction--sampling} paradigm~\cite{ml4co-bench}: a scoring oracle produces a heatmap over variables, and a dedicated sampling procedure then draws from the induced distribution to produce a feasible solution. Such pipelines have been shown to achieve strong results on maximum independent set, maximum cut, and related problems~\cite{li_gnn_tree_search,ahn_lwd_mis,coexpander}. The ML-based approach of Shen et al. is shown to outperform exact solvers and hand-crafted heuristics, making it the state-of-the-art for ML-enhanced pricing. The real contribution of ML to performance of CG and B\&P is the topic of our work, and we use MLPH as the primary baseline.

A growing body of work questions whether the learned components genuinely earn the gains attributed to them. B\"other et al.~\cite{boether-critique} find that the advantages of a GNN-based MIS solver come largely from the classical components around the learned model; Sun et al.~\cite{revisiting_sampling_co} show that non-learned sampling matches ML-based approaches on standard MIS benchmarks, while Santana et al.~\cite{Santana2023} report a related finding for vehicle routing. We argue that the same perspective applies to solving graph coloring via B\&P with ML-enhanced pricing: the RMP dual solution, computed at every CG iteration, already provides a natural non-learned heatmap fitting the same prediction--sampling paradigm that MLPH relies on. This raises the question of how much MLPH's learned scorer adds over information that is already available --- the main question we address in this paper.

\section{Problem Formulation}
\label{sec:problem}

Let $G = (V, E)$ be an undirected graph with vertex set $V$ and edge set $E$. The graph coloring problem (GCP) consists in partitioning the vertex set $V$ into the minimum number of independent sets, also referred to as color classes. Recall that a subset of vertices $S \subseteq V$ is called an independent set if no two vertices in $S$ are adjacent in $G$. Equivalently, the GCP seeks the smallest number of such subsets whose union covers $V$.

A well-known way of encoding the GCP relies on a set covering formulation defined over the family $\mathcal{I}$ of all maximal independent sets of the graph. The corresponding linear programming relaxation of the problem can be written as
\begin{equation}\label{eq:gcp}
    \min\bigl\{\textstyle\sum_{S \in \mathcal{I}} y_S 
    \;:\; 
    \textstyle\sum_{v \in S} y_S \geq 1 \;\; \forall v \in V,\;\;
    y_S \geq 0 \;\; \forall S \in \mathcal{I}
    \bigr\}.
\end{equation}
Here, each variable $y_S$ stands for an independent set $S \in \mathcal{I}$ and indicates whether $S$ participates in the covering of $V$. The constraints ensure that every vertex $v \in V$ is contained in at least one selected independent set.

Since the number of maximal independent sets can grow exponentially with $|V|$, an explicit enumeration of the entire family $\mathcal{I}$ is computationally intractable. For this reason, the linear relaxation~\eqref{eq:gcp} is typically solved by means of column generation. In this approach, the optimization is initially performed on a restricted master problem that involves only a subset $\mathcal{I}' \subseteq \mathcal{I}$ of independent sets, and additional ones are generated dynamically through the iterative solution of a pricing subproblem. By analogy with the way new variables enter the basis when a linear program is solved, the independent sets are called \emph{columns}.

Let $\pi = (\pi_v)_{v \in V} \geq 0$ denote the dual values associated with the constraints~\eqref{eq:gcp} in an optimal solution of the RMP. Then for any independent set $S \subseteq V$, the \emph{reduced cost} of the associated variable $y_S$ is given by
\begin{equation}
\bar{c}_S = 1 - \sum_{v \in S} \pi_v .
\label{eq:reduced_cost}
\end{equation}

The pricing step consists of identifying a column with a minimal reduced cost, which equivalently leads to the formulation of the maximum weight independent set (MWIS) problem:
\begin{equation}\label{eq:mwis}
    \max\bigl\{\textstyle\sum_{v \in V} \pi_v x_v 
    \;:\;
    x_u + x_v \leq 1 \;\; \forall (u,v) \in E,\;\;
    x_v \in \{0,1\} \;\; \forall v \in V
    \bigr\}.
\end{equation}

According to the optimality conditions of linear programming~\cite{BertsimasTsitsiklis1997}, a column corresponding to an independent set $S$ can improve the current RMP solution if and only if its reduced cost~$\bar{c}_S = 1 - \sum_{v \in S} \pi_v$ is negative, equivalently, if and only if $\sum_{v \in S} \pi_v > 1$. If no independent set satisfies this condition, the current RMP solution is optimal for the linear relaxation of the GCP.

To obtain an integer optimal solution, the column generation is embedded within a branch-and-bound framework, yielding a branch-and-price (B\&P) method. At each node of the search tree, the LP relaxation is solved by column generation; its objective value provides a lower bound on the chromatic number. Branching decisions then split the current node into children by imposing additional constraints---for instance, requiring that two vertices do or do not receive the same color---and column generation is restarted in each child under the imposed constraints. Feasible integer colorings discovered during the search provide upper bounds. The algorithm terminates when the lower and upper bounds coincide, proving optimality, or when a predefined computational limit is reached.

\section{Solution Methodology}
\label{sec:methodology}

In this section we first formalize the prediction--sampling paradigm shared between the Dual and MLPH heuristic pricers (Algorithm~\ref{alg:ps_pricing}), and then proceed with the description of every pricing method we compare against, as well as the surrounding CG/B\&P framework.

State-of-the-art heuristic pricing strategies for the GCP follow a two-stage \emph{prediction--sampling} paradigm~\cite{Shen2022}. In the first stage, a scoring procedure assigns each vertex $v$ a score $s_v \in [0,1]$, derived from the graph $G$ and the dual solution $\pi$ of the restricted master problem (RMP), reflecting how likely $v$ is to belong to a high-quality independent set. In the second stage, a sampling procedure uses these scores to construct $K$ candidate independent sets greedily. Given a score vector $s = (s_v)_{v \in V}$, the construction is repeated $K$ times: starting from the candidate set $U = V$ and $S = \emptyset$, vertices are selected iteratively---if the total score of remaining candidates $\sum_{u \in U} s_u$ falls below a numerical-stability threshold $\varepsilon$, a vertex is chosen uniformly at random, otherwise sampled with probability proportional to $s_v$; the chosen vertex is added to $S$, and both it and its neighbours are removed from $U$ to preserve independence. After all $K$ sets are generated, those with negative reduced cost (Eq.~\eqref{eq:reduced_cost}) are added to the RMP. We adopt this paradigm with a fixed sampling procedure, varying only the scoring rule.

\begin{algorithm}[!t]
  \small
  \caption{Prediction--sampling heuristic pricing}
  \label{alg:ps_pricing}
  \begin{algorithmic}[1]
  \REQUIRE Graph $G=(V,E)$, dual solution $\pi$, scoring procedure $\mathcal{O}$, sampling budget $K$, threshold $\varepsilon > 0$
  \ENSURE Set $C$ of independent sets with negative reduced cost
  \STATE Compute scores $s_v \leftarrow \mathcal{O}(G,\pi)$ for all $v \in V$
  \STATE $C \leftarrow \emptyset$
  \FOR{$k = 1$ \TO $K$}
    \STATE $U \leftarrow V$, \quad $S \leftarrow \emptyset$
    \WHILE{$U \neq \emptyset$}
      \IF{$\sum_{u \in U} s_u < \varepsilon$}
        \STATE Choose $v \in U$ uniformly at random
      \ELSE
        \STATE Choose $v \in U$ with probability $s_v / \sum_{u \in U} s_u$
      \ENDIF
      \STATE $S \leftarrow S \cup \{v\}$
      \STATE $U \leftarrow U \setminus \bigl(\{v\} \cup N(v)\bigr)$
    \ENDWHILE
    \IF{$\sum_{v \in S} \pi_v > 1$}
      \STATE $C \leftarrow C \cup \{S\}$
    \ENDIF
  \ENDFOR
  \RETURN $C$
  \end{algorithmic}
\end{algorithm}

\subsection{Pricing Methods}
\label{sec:score_methods}

We consider seven pricing methods organized into three groups. The approaches in the first group---Dual, MLPH, and Random---share the prediction--sampling skeleton of Algorithm~\ref{alg:ps_pricing} and differ only in the scoring rule. Two other approaches---ACO and FastWClq---replace prediction--sampling by a direct search over the dual-weighted graph. The remaining two---ILP and CP---solve the MWIS subproblem~\eqref{eq:mwis} to optimality.

The proposed \emph{Dual} approach uses the RMP dual solution $\pi$ directly as the vertex score vector, $s_v = \pi_v$; since the duals are produced by the RMP solve at every CG iteration anyway, the scoring step adds no extra computation. \emph{MLPH}~\cite{Shen2022} uses a learned scorer: each vertex is described by a feature vector derived from the dual solution and the graph (the dual weight $\pi_v$, the degree $\deg(v)$, a non-neighbourhood bound $\max_{u \notin N(v)} \pi_u$, and rank statistics over random IS samples), and a linear SVM outputs scores $s_v \in [0,1]$ through a logistic function. We use the original C++ implementation of Shen et al., so any runtime difference between MLPH and the other oracles cannot be attributed to a slower or weaker reproduction of the learned model. \emph{Random} assigns the uniform score $s_v \equiv 1/|V|$ to every vertex and uses it directly for sampling, similarly to the Dual approach.

The two methods of this (direct-search) group are included because Shen et al.~\cite{Shen2022} benchmark MLPH against them in the original paper. \emph{ACO}~\cite{Dorigo1996} is a dynamic-pheromone ant-colony metaheuristic in which pheromone trails over vertices are reinforced by every IS produced and decay each iteration; up to $K$ best independent sets along the trajectory are returned. \emph{FastWClq}~\cite{CaiLin2016} applies the Cai--Lin local-search heuristic for the maximum-weight clique problem to the complement graph (since MWIS on $G$ is MaxClique on $\bar{G}$); $K$ restarts with offset seeds yield $K$ distinct candidate independent sets. For both methods we use the C++ implementation bundled with the MLPH release of Shen et al.~\cite{Shen2022}.

We also include two exact MWIS solvers as a reference. \emph{Exact ILP} formulates problem~\eqref{eq:mwis} as an integer linear program in Gurobi and extracts up to $K$ independent sets ranked by most-negative reduced cost in a single run via the solution-pool mechanism. The Gurobi run explicitly enables the standard MIP-heuristics suite (RINS, RENS-style restarts, the feasibility pump, and no-relaxation primal heuristics), so the comparison implicitly covers heuristics of this type as well. \emph{Exact CP} encodes problem~\eqref{eq:mwis} as a feasibility problem in IBM CP Optimizer: vertex variables $x_v \in \{0,1\}$ subject to edge exclusion and a per-call cutoff $\sum_v \pi_v x_v \geq 1 + \varepsilon$; up to $K$ feasible (improving) independent sets are enumerated within a single solver session. The two exact pricers complement each other in the kind of column they prefer: ILP, configured for $K$-best-by-objective solutions, returns the most negative-reduced-cost columns, whereas CP, run as pure satisfaction, returns the first $K$ improving columns it can certify, irrespective of objective margin.

\subsection{Column Generation and Branch-and-Price}
\label{subsec:cg-bp}
At each CG iteration, the chosen pricer is invoked on the current dual solution and adds up to $K$ improving columns to the RMP; if none are returned, a single-column exact MWIS fallback is invoked --- the same Gurobi MWIS formulation as Exact ILP but tuned to stop at the first improving column (no extended MIP-heuristics suite, no $K$-best solution pool) --- and the procedure terminates when both stages fail to find an improving column.
Integer solutions are obtained by wrapping CG in a best-first branch-and-price tree (open nodes ordered by their LP bound) with Ryan--Foster branching~\cite{Ryan1981,MehrotraTrick1996} and a DSATUR heuristic for computing the upper bound~\cite{Brelaz1979}; a Lagrangian fallback bound is used when CG at a node times out. The full pseudocode (Algorithms~\ref{alg:cg} and~\ref{alg:bp}) and the derivation of the Lagrangian bound are given in Appendix~\ref{app:framework}. We also note that across all pricing methods, only the pricer itself varies, while all the other components (branching, upper bounding, pruning, and Lagrangian-bound computation) are identical and deliberately kept at a base level. By foregoing the further refinements surveyed in Section~\ref{sec:related_work} (RMP stabilization, alternative branching schemes, decision-diagram formulations, tighter lower bounds), we ensure that observed performance differences must be attributed to the pricing component alone.

\section{Experimental Evaluation}
\label{sec:experiments}

In this section we evaluate the compared pricing approaches within a unified CG and B\&P framework implemented in Python~3.11 using Gurobi~13 and IBM CP Optimizer.\footnote{The source code and data will be made publicly available after the review process.} We first describe the experimental setup; Section~\ref{subsec:k-sensitivity} then reports an ablation on the sampling budget~$K$ and identifies the value at which each method achieves its best performance. Section~\ref{subsec:cg-performance} compares end-to-end CG and B\&P performance on the benchmark of Shen et al.~\cite{Shen2022}.

\paragraph*{Benchmarks.}\label{subsec:setup}
We use two benchmark families. The first is the Google graph coloring benchmark suite\footnote{\url{https://sites.google.com/site/graphcoloring/vertex-coloring}}, the same dataset used by Shen et al.~\cite{Shen2022}. It contains 81 graphs of diverse sizes and densities and serves as the primary benchmark for our comparison. To study how pricing method performance depends on graph structure, we group the instances into four macro-families: \emph{Recursive sparse}---Mycielski-style constructions (Mycielski, Insertions, FullIns) designed to have high chromatic number despite few edges (23 instances); \emph{DIMACS random}---uniform, geometric, planted-$\chi$, and recursive random graphs from the DSJC, DSJR, flat, and $r$-series of the DIMACS coloring challenge (25 instances); \emph{Structured combinatorial}---graphs from combinatorial constructions with a planned chromatic number (Leighton le450, Queens; 20 instances); and \emph{Real-world}---instances from compiler register allocation (GPIA), school timetabling, wireless access points (WAP), and the MUG family of unsatisfiable colorings (13 instances). The second benchmark is a collection of Erd\H{o}s--R\'{e}nyi random graphs $G(n,p)$ with $n \in \{100, 200, 300, 400, 500\}$ and $p \in \{0.1, 0.3, 0.5, 0.7, 0.9\}$. Each $(n,p)$ pair is instantiated with 10 random seeds, for 250 instances in total. These graphs form a parametric grid over size and density that we use to calibrate the sampling budget~$K$ for each pricing method before evaluating it on the Google benchmark.

\paragraph*{Framework and evaluation procedure.}
All pricing approaches are evaluated within the CG and B\&P framework of Sections~\ref{sec:problem} and~\ref{sec:methodology}. Our setup follows Shen et al.~\cite{Shen2022} with two changes that isolate the effect of the pricing oracle. First, we initialize the RMP with $|V|$ random independent sets (plus singleton columns for any uncovered vertices), rather than $10|V|$, so that the initial RMP does not mask the effect of the pricer. Second, if the exact fallback is triggered after a heuristic pricing failure, it stops at the first improving column to keep its runtime impact small. In our setting, each pricing call is limited to 60\,s. Total time limits are 3\,600\,s for CG and 7\,200\,s for B\&P. The $K$-ablation uses seed~39 with a 600\,s limit per run. Downstream CG and B\&P experiments are repeated over two independent seeds (42--43). All experiments are CPU-only and run on a single Linux server with x86-64 CPUs and at least 64\,GB of RAM.

\subsection{Sensitivity to Sampling Budget \texorpdfstring{$K$}{K}}
\label{subsec:k-sensitivity}

\begin{figure}[t]
  \centering
  \input{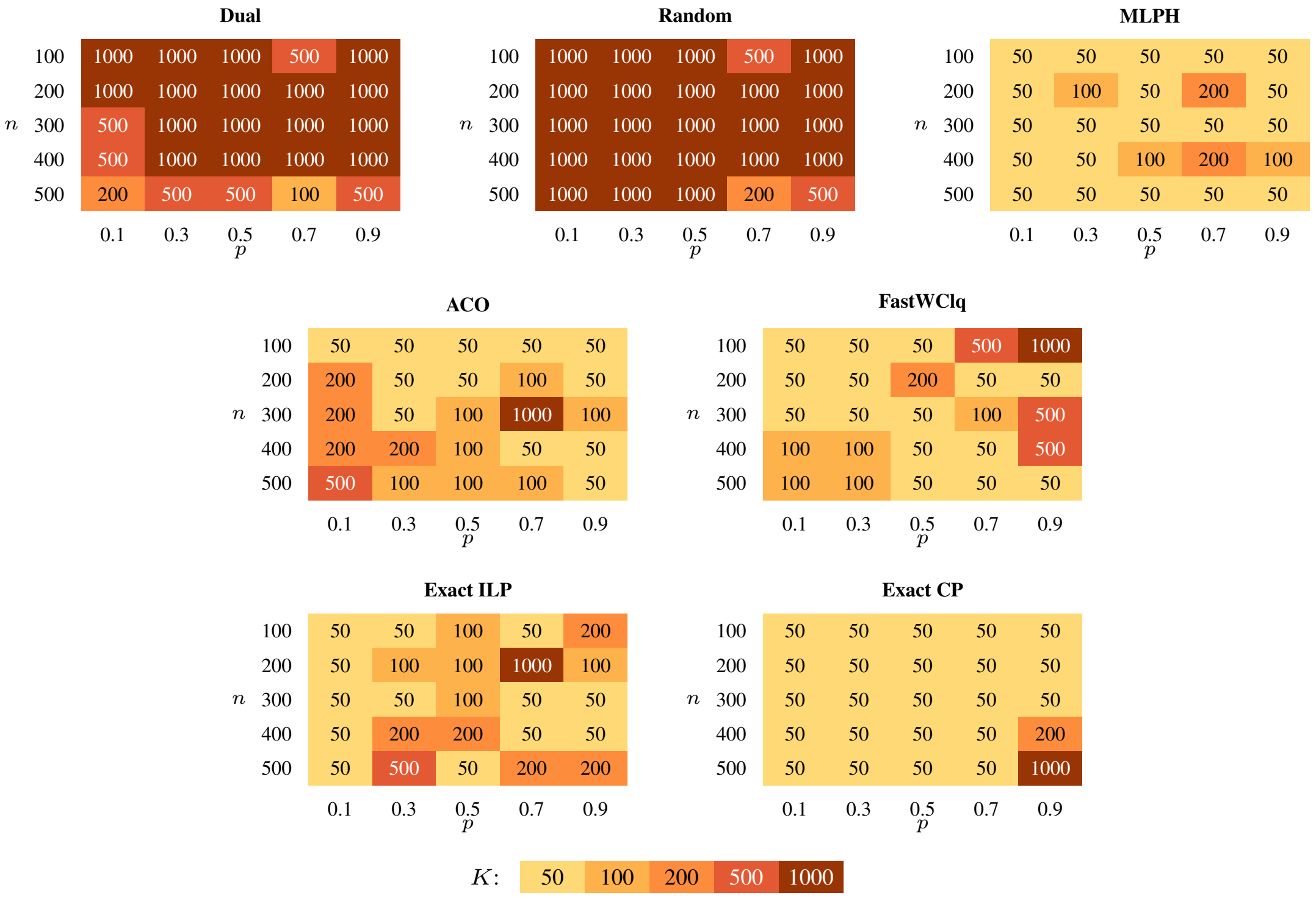}
  \caption{Selected sampling budget $K^{*}$ for each pricer, size, and density.}
  \label{fig:ablation}
\end{figure}

The sampling budget~$K$ controls how many candidate independent sets are generated per pricing call. For the prediction--sampling pricers (Dual, MLPH, Random), $K$ is the number of samples drawn by the sampling procedure. For ACO and FastWClq, $K$ caps the number of independent sets returned by the search. For the exact oracles (ILP, CP), $K$ caps the number of feasible solutions extracted per call. We examine how sensitive each method is to~$K$ and select a value~$K^{*}$ per (method, size, density) triple for use in all subsequent experiments. We use the full Erd\H{o}s--R\'{e}nyi dataset of 250 instances. For each method and each $K \in \{50, 100, 200, 500, 1000\}$, we run CG (without branching) on every instance with a 600\,s time limit, for 8\,750 runs in total. The 600\,s limit keeps the ablation in the heuristic-only regime, so that $K^{*}$ is not affected by exact-fallback calls, which are rare in early iterations. Results are averaged over the 10 instances in each $(n, p)$ cell. The resulting grid (Figure~\ref{fig:ablation}) fixes $K^{*}$ separately for each pricer, size, and density --- these are the values used in all subsequent experiments --- and shows how $K^{*}$ varies across methods and graph categories.

Figure~\ref{fig:ablation} demonstrates that the methods exhibit two types of behavior with respect to their best $K$: Dual and Random prefer the largest budget (mostly $K{=}1000$), while MLPH, ACO, FastWClq, ILP, and CP prefer the smallest ($K{=}50$ dominates). The MLPH--Dual contrast is the most informative one: both methods use the same sampling procedure (Algorithm~\ref{alg:ps_pricing}) and differ only in the score vector fed into it. The small optimal $K$ for MLPH is consistent with the SVM producing a peaked score distribution: once the few high-score vertices have been sampled, more draws yield little diversity. The scores used by Dual are spread more evenly across vertices, so each new sample has in general a higher chance to land in a different region of the graph and larger $K$ keeps adding new columns. Random has the flattest distribution and lands at $K{=}1000$ for the same reason. For the search-based oracles (ACO, FastWClq, ILP, CP), $K$ plays a different role: it caps the number of independent sets returned per call, not the size of a sampling distribution. Their preference for small $K$ is consistent with finding a few good columns early within each call, after which additional search time within the same per-call budget tends to bring little extra value.

\subsection{Column Generation and Branch-and-Price Results}
\label{subsec:cg-performance}

We now compare the pricing approaches on the Google graph coloring benchmark within the CG and B\&P framework. For each benchmark instance, we set the per-method $K$ in three steps: (i) compute the instance's size $n$ and edge density $p$; (ii) find the ER grid cell in $\{100, 200, 300, 400, 500\} \times \{0.1, 0.3, 0.5, 0.7, 0.9\}$ closest to $(n, p)$ in Euclidean distance after normalizing each axis by its grid maximum; (iii) for each pricing method, use the $K^{*}$ that the ablation selected for the resulting (size, density) cell. Table~\ref{tab:results} reports the main CG and B\&P metrics for each pricing method (the upper part covers CG and the lower part covers B\&P). Each cell reports two numbers: the shifted geometric mean (SGM, shift\,=\,1) of the runtime in seconds (SGM is less sensitive to outliers than the arithmetic mean), and the mean objective value (respectively, the final RMP objective for CG and the best integer incumbent for B\&P).

\begin{table}[!t]
  \centering
  \caption{Comparison of pricing methods in CG and B\&P on the Google benchmark instances.}
  \label{tab:results}
  \small
\setlength{\tabcolsep}{4pt}
\begin{tabular}{lrrrrr}
\toprule
Method & Rec.\ sparse & Random & Struct. & Real-world & All \\
\multicolumn{6}{l}{\textit{Each cell: SGM runtime (s) / mean objective.}} \\
\midrule
\multicolumn{6}{l}{\textbf{Column Generation}} \\
\midrule
Dual & \textbf{101} / \textbf{3.7} & \textbf{1062} / \textbf{66.4} & \textbf{473} / \textbf{14.0} & \textbf{412} / \textbf{18.3} & \textbf{384} / \textbf{28.0} \\
Random & 348 / 4.0 & 1258 / 66.8 & 863 / 15.0 & 845 / 22.8 & 747 / 29.1 \\
MLPH & 448 / 4.2 & 1633 / 76.8 & 1032 / 16.1 & 1137 / 26.1 & 953 / 33.0 \\
ACO & 632 / 4.3 & 1643 / 77.2 & 1028 / 16.4 & 1374 / 27.7 & 1084 / 33.5 \\
FastWClq & 512 / 4.2 & 1780 / 77.3 & 1266 / 16.4 & 1374 / 29.6 & 1102 / 33.8 \\
Exact ILP & 986 / 4.1 & 2434 / 74.5 & 1190 / 14.6 & 2249 / 21.6 & 1559 / 31.2 \\
Exact CP & 2181 / 5.6 & 1935 / 80.9 & 1321 / 18.1 & 3400 / 32.3 & 1994 / 36.2 \\
\midrule
\multicolumn{6}{l}{\textbf{Branch-and-Price}} \\
\multicolumn{6}{l}{\textit{Each cell: SGM runtime (s) / mean objective / mean nodes.}} \\
\midrule
Dual & \textbf{5638} / \textbf{6.0} / \textbf{1004} & 5971 / \textbf{79.0} / 27 & \textbf{5720} / 16.8 / 65 & \textbf{953} / \textbf{20.6} / \textbf{4} & \textbf{4330} / \textbf{33.5} / \textbf{310} \\
Random & 5877 / 6.0 / 653 & \textbf{5570} / 82.1 / 24 & 5990 / 17.0 / 67 & 1836 / 20.7 / 3 & 4819 / 34.6 / 210 \\
MLPH & 6078 / 6.0 / 466 & 6272 / 80.6 / 30 & 6316 / 16.8 / 71 & 2095 / 20.7 / 3 & 5222 / 34.1 / 160 \\
ACO & 6004 / 6.0 / 463 & 6494 / 80.5 / \textbf{33} & 6475 / \textbf{16.6} / \textbf{80} & 2442 / 20.6 / 3 & 5424 / 34.0 / 162 \\
FastWClq & 6120 / 6.0 / 394 & 6671 / 80.4 / 32 & 6764 / 16.8 / 55 & 2331 / 20.7 / 2 & 5518 / 34.0 / 136 \\
Exact ILP & 6889 / 6.0 / 116 & 6861 / 82.3 / 12 & 6637 / 16.9 / 45 & 3495 / 20.7 / 2 & 6114 / 34.6 / 48 \\
Exact CP & 7200 / 6.0 / 46 & 6764 / 82.1 / 13 & 7200 / 16.9 / 20 & 6113 / 20.8 / 2 & 6879 / 34.5 / 22 \\
\bottomrule
\end{tabular}

\end{table}

The CG results clearly show that the proposed Dual approach attains the lowest runtime --- from $\sim\!2\times$ to $\sim\!5\times$ faster than the baselines --- and it is also the best method on every macro-family with respect to the mean objective value. Among the rest, Random ranks second despite using no learned or dual information, followed by MLPH and the search-based heuristics (ACO, FastWClq) with comparable runtimes; the exact oracles (ILP, CP) are the slowest, since the per-call cost of solving MWIS to optimality outweighs the benefits of returning multiple improving columns with the highest potential.

\begin{table}[!t]
  \centering
  \caption{Heuristic-success rate per macro-family: the share of CG iterations on which the pricer returns at least one improving column.}
  \label{tab:heur-success}
  \small
\setlength{\tabcolsep}{6pt}
\begin{tabular}{lrrrrr}
\toprule
Method & Recursive sparse & DIMACS random & Structured & Real-world & All \\
\midrule
Dual & \textbf{69.4\%} & \textbf{82.6\%} & \textbf{91.7\%} & \textbf{91.1\%} & \textbf{83.6\%} \\
Random & 37.2\% & 78.1\% & 59.0\% & 83.7\% & 59.4\% \\
MLPH & 10.4\% & 26.5\% & 20.0\% & 21.1\% & 18.4\% \\
ACO & 4.3\% & 9.3\% & 4.4\% & 6.9\% & 5.8\% \\
FastWClq & 6.0\% & 19.9\% & 11.6\% & 16.1\% & 12.4\% \\
\bottomrule
\end{tabular}

\end{table}

Dual also achieves the best aggregate B\&P runtime and, in the per-family breakdown, wins three out of four families both by runtime and by mean objective. Random ranks second, followed by MLPH, ACO, and FastWClq with comparable runtimes; the exact pricing approaches are uniformly the slowest. The CG advantage of Dual carries over to B\&P but in weaker form --- the $\sim\!2$--$5\times$ gap at the LP level shrinks to about $\sim\!1.2$--$1.5\times$ --- because the surrounding pipeline (branching, the random initial column pool, and most importantly the DSATUR upper bound, which efficiently produces good upper bound for most instances independently of the pricer) also contributes. The largest benefit of Dual is on \emph{real-world} instances, where the surrounding pipeline interferes the least with the pricing signal. Two exceptions, where Dual approach does not lead, deserve mention. On the \emph{structured combinatorial} family, ACO achieves a slightly better mean incumbent than Dual and explores more search-tree nodes, having potentially better chances to achieve better solution quality; however, the advantage is moderate. On the \emph{random} family, Random approach achieves a lower mean runtime by solving the smaller instances faster, yet Dual still produces the lowest mean objective on this family. Notably, the B\&P performance reflects variability introduced by the surrounding pipeline rather than purely the properties of the pricer itself --- which is exactly why we report both CG and B\&P numbers: CG isolates the pricing component cleanly, while B\&P combines it with downstream pipeline noise.

A complementary analysis is presented in Table~\ref{tab:heur-success}, reporting the share of CG iterations on which the pricer returns at least one improving column without falling back to the exact MWIS solver. The purpose is to assess the efficiency of each pricing method in generating useful columns. As shown in the table, Dual achieves the highest rate on every family; Random ranks second everywhere, substantially above MLPH. Two points are worth noting about this analysis. First, the metric counts whether \emph{any} improving column is returned, not the magnitude of its reduced cost: MLPH can concentrate on a few high-value columns early and let the master converge before pricers with more diffuse scores yield equivalent progress. Second, ACO and FastWClq succeed least often (5--12\%), yet ACO produces the best B\&P incumbent on the structured combinatorial family (Table~\ref{tab:results}, lower section)---per-iteration pricing efficiency therefore does not always translate directly into B\&P solution quality, reinforcing the framework-level point that B\&P numbers reflect pricing \emph{plus} downstream pipeline noise rather than pricing alone.

\subsection{Dual vs.\ MLPH across graph families}
\label{subsec:family-analysis}

\begin{figure}[!t]
  \centering
  \begin{minipage}[t]{0.48\linewidth}
    \centering
    \begin{tikzpicture}
\definecolor{dualcolor}{RGB}{31,119,180}
\definecolor{mlphcolor}{RGB}{255,127,14}
\begin{axis}[
    width=7cm, height=5cm, ymode=log,
    ybar=2pt, bar width=8pt,
    symbolic x coords={Recursive sparse, DIMACS random, Structured, Real-world},
    xtick=data,
    xticklabel style={font=\scriptsize, rotate=20, anchor=north east},
    yticklabel style={font=\scriptsize},
    ylabel={CG SGM runtime (s)},
    label style={font=\scriptsize},
    enlarge x limits=0.20,
    legend style={font=\scriptsize, at={(0.5,1.02)}, anchor=south, legend columns=2, draw=none},
    legend cell align=left,
    ymin=0.5,
]
\addplot+[fill=dualcolor, draw=dualcolor!70!black] coordinates {(Recursive sparse,101.4434) (DIMACS random,1063.0472) (Structured,473.4050) (Real-world,411.7986)};
\addlegendentry{Dual}
\addplot+[fill=mlphcolor, draw=mlphcolor!70!black] coordinates {(Recursive sparse,448.0726) (DIMACS random,1633.1685) (Structured,1032.4970) (Real-world,1137.2965)};
\addlegendentry{MLPH}
\end{axis}
\end{tikzpicture}
  \end{minipage}\hfill
  \begin{minipage}[t]{0.48\linewidth}
    \centering
    \begin{tikzpicture}
\definecolor{famRecursivesparse}{RGB}{31,119,180}
\definecolor{famDIMACSrandom}{RGB}{255,127,14}
\definecolor{famStructured}{RGB}{44,160,44}
\definecolor{famRealworld}{RGB}{214,39,40}
\begin{axis}[
    width=7cm, height=5cm,
    xmode=log, ymode=log,
    xmin=1, xmax=4000, ymin=1, ymax=4000,
    xlabel={MLPH runtime (s)},
    ylabel={Dual runtime (s)},
    tick label style={font=\scriptsize},
    label style={font=\scriptsize},
    legend style={font=\scriptsize, at={(0.02,0.98)}, anchor=north west, draw=none, fill=white, fill opacity=0.8, text opacity=1},
    legend cell align=left,
    grid=both, grid style={very thin, gray!20},
]
\addplot[domain=1:4000, samples=2, dashed, gray, forget plot] {x};
\addplot[only marks, mark=*, mark size=1.6pt, color=famRecursivesparse] coordinates {(309.10084850000004,1.7468075) (1818.3023755,113.999837) (101.960469,9.109875500000001) (3600.0,3600.0) (3600.0,3600.0) (1825.383625,17.424553) (3600.0,177.4528175) (9.450839,1.620024) (3600.0,3430.1744685) (3600.0,3600.0) (1836.315705,19.981537000000003) (3600.0,680.1817265) (37.934805999999995,8.9188455) (3600.0,3600.0) (3600.0,3600.0) (1918.6518585000001,53.543413) (197.40097200000002,57.642830000000004) (3600.0,3600.0) (2592.9792509999997,129.1015065) (0.6615645,0.19992500000000002) (18.337,1.5232705) (570.6418035,7.516215000000001) (2025.8703005,56.011833499999995)};
\addlegendentry{Recursive sparse}
\addplot[only marks, mark=square*, mark size=1.6pt, color=famDIMACSrandom] coordinates {(3600.0,3600.0) (3600.0,3600.0) (3600.0,3600.0) (3600.0,3469.3733695) (1112.688535,174.2860595) (84.82160400000001,8.888709500000001) (3600.0,3600.0) (3600.0,3600.0) (1884.6480935,402.5384935) (3600.0,3600.0) (3600.0,3600.0) (2959.3870644999997,3600.0) (1823.9254274999998,28.071740499999997) (2002.5211695,292.6735725) (3600.0,3600.0) (3600.0,3600.0) (3600.0,3600.0) (3600.0,3600.0) (3600.0,3600.0) (3600.0,3600.0) (1928.0510495,256.23767100000003) (2473.2544005,1195.9703665) (3600.0,3600.0) (51.8056715,44.074914500000006) (1888.0083875,74.0587015)};
\addlegendentry{DIMACS random}
\addplot[only marks, mark=triangle*, mark size=1.6pt, color=famStructured] coordinates {(3600.0,1117.8221545) (2835.9220975,392.6483365) (3600.0,3600.0) (3600.0,3600.0) (1119.291227,22.579349999999998) (3600.0,3600.0) (3600.0,3600.0) (3600.0,3600.0) (3600.0,3600.0) (3600.0,3600.0) (3600.0,3600.0) (311.9369895,48.196422) (677.0479465,71.3570945) (1341.335176,101.39659449999999) (1904.8530315,102.3051815) (1954.9627095,486.5192705) (2030.7720065,780.1834855) (1988.0742165,363.2151695) (40.597108,9.294227) (145.49443549999998,20.282080999999998)};
\addlegendentry{Structured}
\addplot[only marks, mark=diamond*, mark size=1.6pt, color=famRealworld] coordinates {(3600.0,3600.0) (1839.8027145,66.3295425) (468.0895365,60.611184) (408.91144699999995,65.455977) (214.41436050000002,28.7752515) (182.3901105,31.406264) (3600.0,3600.0) (3600.0,3600.0) (1934.3313125,154.3615325) (3600.0,2343.2024149999997) (3600.0,3600.0) (3600.0,3600.0) (1962.488626,182.10397949999998)};
\addlegendentry{Real-world}
\end{axis}
\end{tikzpicture}
  \end{minipage}
  \caption{\textbf{(left)}~CG runtime per family for Dual and MLPH. \textbf{(right)}~Per-instance scatter of Dual vs.\ MLPH CG runtime (log scale; dashed line is $y{=}x$); points below the diagonal mark instances where Dual converges faster than MLPH.}
  \label{fig:dual-vs-mlph}
\end{figure}
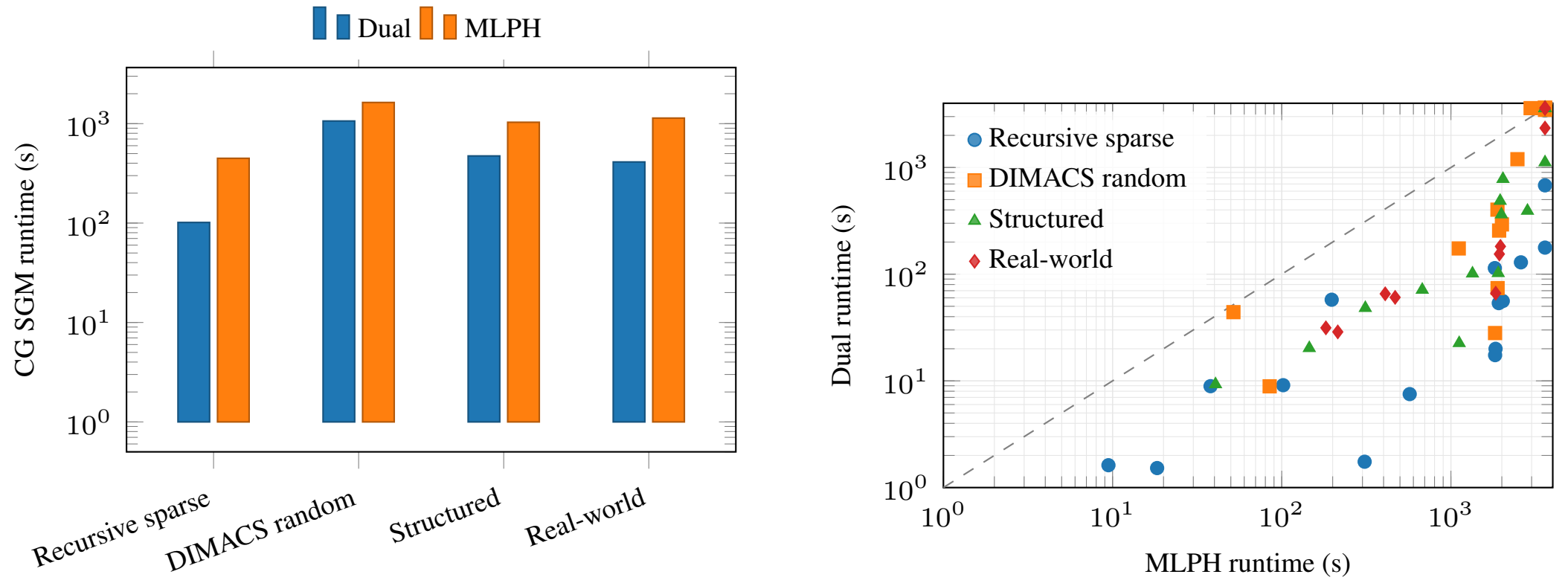

Table~\ref{tab:results} already shows the per-family aggregates for all seven oracles; here we zoom in on the Dual~vs.~MLPH comparison, which is the central methodological question of the paper. Table~\ref{tab:results-per-family} reports four per-family metrics side by side: \emph{Solved} --- the share of CG runs that converge within the time limit; \emph{Heur.\ success} --- the heuristic-success rate (defined for Table~\ref{tab:heur-success}); and SGM runtimes for CG and B\&P. Figure~\ref{fig:dual-vs-mlph} provides the corresponding visualization.

\begin{table}[!t]
  \centering
  \caption{Per-family Dual vs.\ MLPH comparison on the Google benchmark. Bold marks the better value within each pair.}
  \label{tab:results-per-family}
  \small
\setlength{\tabcolsep}{4pt}
\begin{tabular}{l|rr|rr|rr|rr}
\toprule
 & \multicolumn{2}{c|}{Solved (\%)} & \multicolumn{2}{c|}{Heur. success (\%)} & \multicolumn{2}{c|}{CG SGM (s)} & \multicolumn{2}{c}{B\&P SGM (s)} \\
Family & Dual & MLPH & Dual & MLPH & Dual & MLPH & Dual & MLPH \\
\midrule
\textbf{Rec.\ sparse} & \textbf{71.7\%} & 47.8\% & \textbf{69.4\%} & 10.4\% & \textbf{101.4} & 448.0 & \textbf{5638.1} & 6077.6 \\
\textbf{Random} & \textbf{38.0\%} & 26.0\% & \textbf{82.6\%} & 26.5\% & \textbf{1062.4} & 1632.9 & \textbf{5971.2} & 6271.7 \\
\textbf{Struct.} & \textbf{60.0\%} & 42.5\% & \textbf{91.7\%} & 20.0\% & \textbf{473.4} & 1032.5 & \textbf{5719.8} & 6316.5 \\
\textbf{Real-world} & \textbf{61.5\%} & 42.3\% & \textbf{91.1\%} & 21.1\% & \textbf{411.7} & 1137.0 & \textbf{952.8} & 2095.4 \\
\midrule
\textbf{All} & \textbf{56.8\%} & 38.9\% & \textbf{83.6\%} & 18.4\% & \textbf{384.1} & 953.1 & \textbf{4329.9} & 5222.4 \\
\bottomrule
\end{tabular}

\end{table}

As Table~\ref{tab:results-per-family} shows, Dual outperforms MLPH on every family. The largest gain is on the \emph{recursive-sparse} family (over $4\times$ CG speedup) and on the \emph{real-world} instances ($\sim\!3\times$), where the Mycielski/Insertions graphs and the register-allocation/timetabling instances are structurally very different from the random graphs Shen et al.\ used to train MLPH. The gap shrinks to about $2\times$ on the \emph{structured combinatorial} family and to $\sim\!1.5\times$ on the \emph{DIMACS random} family, where MLPH was reported strongest in the original evaluation~\cite{Shen2022}, but Dual still leads.

\section{Conclusion}
\label{sec:conclusion}

In this paper, we focused on the role of machine learning in heuristic pricing for branch-and-price applied to graph coloring. The graph coloring problem is NP-hard, and branch-and-price over its set-covering formulation is among the most effective exact approaches to it. Each iteration of column generation calls a pricing subproblem---a maximum weight independent set under the current dual prices---which is itself NP-hard and is typically the runtime bottleneck. The state-of-the-art heuristic response is MLPH~\cite{Shen2022}, which follows a prediction--sampling paradigm: a learned SVM scores each vertex before a randomized sampling procedure draws candidate independent sets. Starting from the observation that this paradigm requires a scoring rule but not necessarily a \emph{learned} one, we proposed \emph{Dual-Weighted Sampling}: feeding the RMP dual solution~$\pi$ directly into the sampling procedure as the per-vertex score vector, bypassing both the learned model and its feature engineering.

Within a unified CG and B\&P framework that holds all components except pricing fixed, Dual-Weighted Sampling outperforms MLPH, the exact ILP and CP MWIS solvers, and the combinatorial heuristics ACO and FastWClq on the Google graph coloring benchmark. The lead over MLPH is largest on graph families structurally distant from the random graphs used to train it (recursive-sparse and real-world instances). Methodologically, this result challenges a common assumption: that capturing useful structure in the pricing step requires a learned model. In MLPH, the master-problem dual solution is the central feature of the learned scorer---and one that is already informative on its own. Adding a learned mapping on top therefore does not turn a weak input into a strong one; it can instead distort the existing information through distribution shift. Whenever such a dual is available, training-free dual scoring should be tried as a baseline before more complex learning pipelines are pursued.

\bibliographystyle{plainnat}
\bibliography{references}

@article{Bengio2021,
  title = {Machine learning for combinatorial optimization: A methodological tour d'horizon},
  volume = {290},
  ISSN = {0377-2217},
  url = {http://dx.doi.org/10.1016/j.ejor.2020.07.063},
  DOI = {10.1016/j.ejor.2020.07.063},
  number = {2},
  journal = {European Journal of Operational Research},
  publisher = {Elsevier BV},
  author = {Bengio, Yoshua and Lodi, Andrea and Prouvost, Antoine},
  year = {2021},
  month = apr,
  pages = {405--421}
}

@book{Lewis2021,
  title     = {Guide to Graph Colouring: Algorithms and Applications},
  ISBN      = {9783030810542},
  ISSN      = {1868-095X},
  url       = {http://dx.doi.org/10.1007/978-3-030-81054-2},
  DOI       = {10.1007/978-3-030-81054-2},
  publisher = {Springer International Publishing},
  author    = {Lewis, R. M. R.},
  year      = {2021},
  series    = {Texts in Computer Science}
}

@article{MehrotraTrick1996,
  author  = {Mehrotra, Anuj and Trick, Michael A.},
  title   = {A Column Generation Approach for Graph Coloring},
  journal = {INFORMS Journal on Computing},
  volume  = {8},
  number  = {4},
  pages   = {344--354},
  year    = {1996},
  doi     = {10.1287/ijoc.8.4.344}
}

@article{Brelaz1979,
  author  = {Br{\'e}laz, Daniel},
  title   = {New Methods to Color the Vertices of a Graph},
  journal = {Communications of the ACM},
  volume  = {22},
  number  = {4},
  pages   = {251--256},
  year    = {1979},
  doi     = {10.1145/359094.359101}
}

@article{GualandiMalucelli2012,
  author  = {Gualandi, Stefano and Malucelli, Federico},
  title   = {Exact Solution of Graph Coloring Problems via Constraint Programming and Column Generation},
  journal = {INFORMS Journal on Computing},
  volume  = {24},
  number  = {1},
  pages   = {81--100},
  year    = {2012},
  doi     = {10.1287/ijoc.1100.0436}
}

@article{DasAhmad2019,
  author  = {Das, Dibyendu and Ahmad, Shahid Asghar and Venkataramanan, Kumar},
  title   = {Deep Learning-based Hybrid Graph-Coloring Algorithm for Register Allocation},
  journal = {CoRR},
  volume  = {abs/1912.03700},
  year    = {2019},
  url     = {https://arxiv.org/abs/1912.03700}
}

@article{CumminsVeras2024,
  author  = {Cummins, Chase and Veras, Richard},
  title   = {Reinforcement Learning for Graph Coloring: Understanding the Power and Limits of Non-Label Invariant Representations},
  journal = {CoRR},
  volume  = {abs/2401.12470},
  year    = {2024},
  url     = {https://arxiv.org/abs/2401.12470}
}

@inproceedings{GasseNeurIPS2019,
  author    = {Gasse, Maxime and Ch{\'e}telat, Didier and Ferroni, Nicola and Charlin, Laurent and Lodi, Andrea},
  title     = {Exact Combinatorial Optimization with Graph Convolutional Neural Networks},
  booktitle = {Advances in Neural Information Processing Systems},
  volume    = {32},
  year      = {2019},
  url       = {https://arxiv.org/abs/1906.01629}
}

@inproceedings{SunErnst2023,
  author    = {Sun, Yuan and Ernst, Andreas T. and Li, Xiaodong and Weiner, Jake},
  title     = {Learning to Generate Columns with Application to Vertex Coloring},
  booktitle = {International Conference on Learning Representations},
  year      = {2023},
  url       = {https://openreview.net/forum?id=JHW30A4DXtO}
}

@article{ml4co-bench,
  title   = {{ML4CO-Bench-101}: Benchmark Machine Learning for Classic Combinatorial Problems on Graphs},
  author  = {Ma, Jianhao and Pan, Zhewei and Li, Yujia and Yan, Junchi and Li, Chao},
  journal = {Advances in Neural Information Processing Systems},
  year    = {2024},
  note    = {Datasets and Benchmarks Track}
}

@inproceedings{coexpander,
  title     = {{COExpander}: Adaptive Solution Expansion for Combinatorial Optimization with Neural Guidance},
  author    = {Sun, Yiding and Li, Kai and Jiang, Chengrun and others},
  booktitle = {Proceedings of the 41st International Conference on Machine Learning},
  year      = {2024}
}

@inproceedings{revisiting_sampling_co,
  title     = {Revisiting Sampling for Combinatorial Optimization},
  author    = {Sun, Haoran and Goshvadi, Katayoon and Nova, Azade and Schuurmans, Dale and Dai, Hanjun},
  booktitle = {Proceedings of the 40th International Conference on Machine Learning},
  year      = {2023}
}

@inproceedings{li_gnn_tree_search,
  title     = {Combinatorial Optimization with Graph Convolutional Networks and Guided Tree Search},
  author    = {Li, Zhenqiu and Chen, Qi and Koltun, Vladlen},
  booktitle = {Advances in Neural Information Processing Systems},
  year      = {2018}
}

@inproceedings{ahn_lwd_mis,
  title     = {Learning What to Defer for Maximum Independent Sets},
  author    = {Ahn, Sungsoo and Shin, Jaehyung and Hottung, Andre and Tierney, Kevin and others},
  booktitle = {Proceedings of the 37th International Conference on Machine Learning},
  year      = {2020}
}

@inproceedings{boether-critique,
  title     = {What Is Wrong with Deep Learning in Tree Search for the Maximum Independent Set Problem?},
  author    = {B{\"o}ther, Henning and Lamm, Martin and Schmidt, Marc and Skutella, Martin},
  booktitle = {Proceedings of the International Conference on Learning Representations},
  year      = {2022}
}

@article{Shen2022,
  title = {Enhancing Column Generation by a Machine-Learning-Based Pricing Heuristic for Graph Coloring},
  volume = {36},
  ISSN = {2159-5399},
  DOI = {10.1609/aaai.v36i9.21230},
  number = {9},
  journal = {Proceedings of the AAAI Conference on Artificial Intelligence},
  publisher = {Association for the Advancement of Artificial Intelligence (AAAI)},
  author = {Shen,  Yunzhuang and Sun,  Yuan and Li,  Xiaodong and Eberhard,  Andrew and Ernst,  Andreas},
  year = {2022},
  month = jun,
  pages = {9926--9934}
}

@article{Malaguti2011,
  title = {An exact approach for the Vertex Coloring Problem},
  volume = {8},
  ISSN = {1572-5286},
  url = {http://dx.doi.org/10.1016/j.disopt.2010.07.005},
  DOI = {10.1016/j.disopt.2010.07.005},
  number = {2},
  journal = {Discrete Optimization},
  publisher = {Elsevier BV},
  author = {Malaguti,  Enrico and Monaci,  Michele and Toth,  Paolo},
  year = {2011},
  month = may,
  pages = {174--190}
}

@article{Held2012,
  title = {Maximum-weight stable sets and safe lower bounds for graph coloring},
  volume = {4},
  ISSN = {1867-2957},
  url = {http://dx.doi.org/10.1007/s12532-012-0042-3},
  DOI = {10.1007/s12532-012-0042-3},
  number = {4},
  journal = {Mathematical Programming Computation},
  publisher = {Springer Science and Business Media LLC},
  author = {Held,  Stephan and Cook,  William and Sewell,  Edward C.},
  year = {2012},
  month = may,
  pages = {363--381}
}

@article{Morrison2014,
  title = {A Wide Branching Strategy for the Graph Coloring Problem},
  volume = {26},
  ISSN = {1526-5528},
  url = {http://dx.doi.org/10.1287/ijoc.2014.0593},
  DOI = {10.1287/ijoc.2014.0593},
  number = {4},
  journal = {INFORMS Journal on Computing},
  publisher = {Institute for Operations Research and the Management Sciences (INFORMS)},
  author = {Morrison,  David R. and Sauppe,  Jason J. and Sewell,  Edward C. and Jacobson,  Sheldon H.},
  year = {2014},
  month = nov,
  pages = {704--717}
}

@article{Ryan1981,
  author  = {Ryan, David M. and Foster, Brian A.},
  title   = {An Integer Programming Approach to Scheduling},
  journal = {Computer Scheduling of Public Transport},
  pages   = {269--280},
  year    = {1981},
  publisher = {North-Holland}
}

@InProceedings{Santana2023,
  author    = {Santana, {\'I}talo and Lodi, Andrea and Vidal, Thibaut},
  editor    = {Cire, Andre A.},
  title     = {Neural Networks for Local Search and Crossover in Vehicle Routing: A Possible Overkill?},
  booktitle = {Integration of Constraint Programming, Artificial Intelligence, and Operations Research},
  year      = {2023},
  publisher = {Springer Nature Switzerland},
  address   = {Cham},
  pages     = {184--199},
  isbn      = {978-3-031-33271-5}
}

@article{LubbeckeDesrosiers2005,
  author  = {L{\"u}bbecke, Marco E. and Desrosiers, Jacques},
  title   = {Selected Topics in Column Generation},
  journal = {Operations Research},
  volume  = {53},
  number  = {6},
  pages   = {1007--1023},
  year    = {2005},
  doi     = {10.1287/opre.1050.0234}
}

@article{Barnhart1998,
  author  = {Barnhart, Cynthia and Johnson, Ellis L. and Nemhauser, George L. and Savelsbergh, Martin W. P. and Vance, Pamela H.},
  title   = {Branch-and-Price: Column Generation for Solving Huge Integer Programs},
  journal = {Operations Research},
  volume  = {46},
  number  = {3},
  pages   = {316--329},
  year    = {1998},
  doi     = {10.1287/opre.46.3.316}
}

@article{Morabit2021,
  author  = {Morabit, Mouad and Desaulniers, Guy and Lodi, Andrea},
  title   = {Machine-Learning-Based Column Selection for Column Generation},
  journal = {Transportation Science},
  volume  = {55},
  number  = {4},
  pages   = {815--831},
  year    = {2021},
  doi     = {10.1287/trsc.2021.1045}
}

@inproceedings{ChiNeurIPS2022,
  author    = {Chi, Cheng and Aboussalah, Amine Mohamed and Khalil, Elias B. and Wang, Juyoung and Sherkat-Masoumi, Zoha},
  title     = {A Deep Reinforcement Learning Framework for Column Generation},
  booktitle = {Advances in Neural Information Processing Systems},
  volume    = {35},
  year      = {2022}
}

@book{BertsimasTsitsiklis1997,
  author    = {Bertsimas, Dimitris and Tsitsiklis, John N.},
  title     = {Introduction to Linear Optimization},
  publisher = {Athena Scientific},
  address   = {Belmont, MA},
  year      = {1997}
}

@article{Vclavk2018,
  title     = {Accelerating the Branch-and-Price Algorithm Using Machine Learning},
  volume    = {271},
  ISSN      = {0377-2217},
  url       = {http://dx.doi.org/10.1016/j.ejor.2018.05.046},
  DOI       = {10.1016/j.ejor.2018.05.046},
  number    = {3},
  journal   = {European Journal of Operational Research},
  publisher = {Elsevier BV},
  author    = {V{\'a}clav{\'\i}k, Roman and Nov{\'a}k, Anton{\'\i}n and {\v{S}}{\r{u}}cha, P{\v{r}}emysl and Hanz{\'a}lek, Zden{\v{e}}k},
  year      = {2018},
  month     = dec,
  pages     = {1055--1069}
}

@article{Kouteck2025,
  title     = {A machine learning approach to rank pricing problems in branch-and-price},
  volume    = {320},
  ISSN      = {0377-2217},
  url       = {http://dx.doi.org/10.1016/j.ejor.2024.07.029},
  DOI       = {10.1016/j.ejor.2024.07.029},
  number    = {2},
  journal   = {European Journal of Operational Research},
  publisher = {Elsevier BV},
  author    = {Kouteck{\'a}, Pavl{\'\i}na and {\v{S}}{\r{u}}cha, P{\v{r}}emysl and H{\r{u}}la, Jan and Maenhout, Broos},
  year      = {2025},
  month     = jan,
  pages     = {328--342}
}

@inproceedings{CaiLin2016,
  author    = {Cai, Shaowei and Lin, Jinkun},
  title     = {Fast Solving Maximum Weight Clique Problem in Massive Graphs},
  booktitle = {Proceedings of the 25th International Joint Conference on Artificial Intelligence (IJCAI)},
  year      = {2016},
  pages     = {568--574}
}

@article{Dorigo1996,
  author    = {Dorigo, Marco and Maniezzo, Vittorio and Colorni, Alberto},
  title     = {Ant System: Optimization by a Colony of Cooperating Agents},
  journal   = {IEEE Transactions on Systems, Man, and Cybernetics, Part B (Cybernetics)},
  volume    = {26},
  number    = {1},
  pages     = {29--41},
  year      = {1996},
  doi       = {10.1109/3477.484436}
}

\newpage
\appendix
\section{CG and B\&P framework details}
\label{app:framework}

This appendix provides the full description of the column-generation and branch-and-price scaffold used in our experiments; the main text gives only a high-level outline.

\paragraph*{CG iteration.}
Let $\mathcal{P}$ denote the current column pool of the RMP. At the root node, $\mathcal{P}$ is initialised with $|V|$ randomly sampled independent sets, with singleton columns added for any uncovered vertices to ensure feasibility; at non-root nodes, $\mathcal{P}$ is inherited from the parent and filtered to remove columns violating the active branching constraints. At each CG iteration, the chosen pricer is invoked: prediction--sampling pricers run Algorithm~\ref{alg:ps_pricing}, ACO/FastWClq run their respective combinatorial searches, and the exact pricers (ILP, CP) enumerate up to $K$ improving independent sets within a single Gurobi or CP Optimizer solver session. If no improving column is found, a single-column exact MWIS solver is called as a fallback. Improving columns are added to $\mathcal{P}$ and the RMP is re-solved; the procedure terminates when neither the chosen pricer nor the fallback finds an improving column. The complete algorithm is given in Algorithm~\ref{alg:cg}.

\begin{algorithm}[!t]
  \small
  \caption{Column generation for GCP}
  \label{alg:cg}
  \begin{algorithmic}[1]
  \REQUIRE Graph $G=(V,E)$, initial column pool $\mathcal{P}_0$, pricing method $M$, sampling budget $K$, threshold $\varepsilon$, time limit $T$
  \ENSURE Final RMP objective $z^*_{\mathrm{LP}}$ and dual vector $\pi$
  \STATE $\mathcal{P} \leftarrow \mathcal{P}_0$
  \STATE $(z^*_{\mathrm{LP}},\, \pi) \leftarrow \textsc{SolveRMP}(\mathcal{P})$
  \WHILE{elapsed time $< T$}
    \STATE $C_1 \leftarrow \textsc{Price}(G, \pi, M, K, \varepsilon)$
    \hfill\COMMENT{LP/MLPH: Alg.~\ref{alg:ps_pricing}; ILP/CP: exact multi-column solver}
    \IF{$C_1 = \emptyset$}
      \STATE $C_2 \leftarrow \textsc{ExactMWIS}(G, \pi)$
      \hfill\COMMENT{single-column exact fallback}
      \IF{$C_2 = \emptyset$}
        \STATE \textbf{break}
        \hfill\COMMENT{no improving column found}
      \ENDIF
      \STATE $\mathcal{P} \leftarrow \mathcal{P} \cup C_2$
    \ELSE
      \STATE $\mathcal{P} \leftarrow \mathcal{P} \cup C_1$
    \ENDIF
    \STATE $(z^*_{\mathrm{LP}},\, \pi) \leftarrow \textsc{SolveRMP}(\mathcal{P})$
  \ENDWHILE
  \RETURN $(z^*_{\mathrm{LP}},\, \pi)$
  \end{algorithmic}
\end{algorithm}

\paragraph*{B\&P scaffold.}
Optimal integer solutions of the GCP are obtained by embedding the column generation procedure within a branch-and-price framework. The tree is explored in best-first order: open nodes are kept in a priority queue keyed by the LP bound at the parent, and the node with the smallest bound is dequeued first. The search runs subject to a time limit~$T$ and a node limit~$N_{\max}$. Branching follows the Ryan--Foster rule~\cite{Ryan1981,MehrotraTrick1996}: when the LP relaxation at a node yields a non-integer solution, the column with value closest to~$0.5$ is selected and a non-adjacent vertex pair $(i,j)$ in it spawns two children---a same-color child (vertices merged via Union--Find, columns containing exactly one of them removed) and a diff-color child (edge $(i,j)$ added to $E$). Upper bounds are obtained by applying the DSATUR greedy heuristic~\cite{Brelaz1979} to the contracted graph at every node; this is independent of the pricer. A node is pruned when $\lceil z_{\mathrm{LP}}(n) \rceil \geq z^*$ (since the chromatic number is integer-valued) or when its LP relaxation is infeasible. When CG at a node does not converge within the time limit, the LP bound is replaced by the Lagrangian bound $z_{\mathrm{LB}}(n) = (\sum_{v \in V} \pi_v) / \max(1, u^*)$, where $\pi$ is the last dual solution and $u^*$ is the LP-MWIS optimum under the current branching constraints; the bound is valid because $u^* \geq \sum_{v \in S} \pi_v$ for every feasible IS~$S$. The complete branch-and-price algorithm is given in Algorithm~\ref{alg:bp}.

\begin{algorithm}[!t]
  \small
  \caption{Branch-and-price for GCP}
  \label{alg:bp}
  \begin{algorithmic}[1]
  \REQUIRE Graph $G=(V,E)$, pricing method $M$, time limit $T$, node limit $N_{\max}$
  \ENSURE Best feasible coloring value $z^*$
  \STATE $z^* \leftarrow \textsc{DSATUR}(G)$
  \hfill\COMMENT{initial upper bound}
  \STATE $\mathcal{P}_r \leftarrow$ root initial column pool
  \STATE $r \leftarrow \textsc{CreateNode}(\emptyset)$;\quad $(z_{\mathrm{LP}}(r),\, \lambda^*(r)) \leftarrow \textsc{CG}(r, \mathcal{P}_r, M)$
  \hfill\COMMENT{Algorithm~\ref{alg:cg}}
  \IF{$\lambda^*(r) \in \{0,1\}^{|\mathcal{P}_r|}$}
    \RETURN $z_{\mathrm{LP}}(r)$
  \ENDIF
  \STATE $\mathcal{Q} \leftarrow \{r\}$;\quad $n_{\mathrm{exp}} \leftarrow 0$
  \hfill\COMMENT{priority queue keyed by $z_{\mathrm{LP}}$}
  \WHILE{$\mathcal{Q} \neq \emptyset$ \AND elapsed time $< T$ \AND $n_{\mathrm{exp}} < N_{\max}$}
    \STATE $n \leftarrow \textsc{PopMin}(\mathcal{Q})$;\quad $n_{\mathrm{exp}} \leftarrow n_{\mathrm{exp}} + 1$
    \hfill\COMMENT{open node with smallest LP bound}
    \IF{$\lceil z_{\mathrm{LP}}(n) \rceil \geq z^*$}
      \STATE \textbf{continue}
      \hfill\COMMENT{prune by bound}
    \ENDIF
    \STATE $(i,j) \leftarrow \textsc{RyanFosterPair}(n, \lambda^*(n))$
    \hfill\COMMENT{most fractional column}
    \FOR{$b \in \{\textsc{Same}(i,j),\; \textsc{Diff}(i,j)\}$}
      \STATE $c \leftarrow \textsc{CreateNode}(n.\mathrm{constraints} \cup \{b\})$
      \STATE $\mathcal{P}_c \leftarrow$ columns inherited from $n$ and filtered by $b$
      \STATE $(z_{\mathrm{LP}}(c),\, \lambda^*(c)) \leftarrow \textsc{CG}(c, \mathcal{P}_c, M)$
      \STATE $z^* \leftarrow \min\bigl(z^*,\; \textsc{DSATUR}(G / \text{constraints}(c))\bigr)$
      \IF{$\lambda^*(c) \in \{0,1\}^{|\mathcal{P}_c|}$ \AND $z_{\mathrm{LP}}(c) < z^*$}
        \STATE $z^* \leftarrow z_{\mathrm{LP}}(c)$
      \ELSIF{$\lceil z_{\mathrm{LP}}(c) \rceil < z^*$}
        \STATE $\textsc{Push}(\mathcal{Q},\, c)$
      \ENDIF
    \ENDFOR
  \ENDWHILE
  \RETURN $z^*$
  \end{algorithmic}
\end{algorithm}

\end{document}